\documentclass[runningheads]{llncs}
\usepackage[T1]{fontenc}

\usepackage{graphicx}
\usepackage{dingbat}
\usepackage{multirow}
\usepackage{
    bbding}
\usepackage[marginal]{footmisc}

\usepackage{booktabs} 

\begin{document}
\title{P2M2-Net: Part-Aware Prompt-Guided Multimodal Point Cloud Completion}
%
%
\author{Linlian Jiang\inst{1} \and
Pan Chen\inst{1} \and
Ye Wang\inst{1} \and
Tieru Wu\inst{1,2} \and
Rui Ma\inst{1,2,\thanks{Rui Ma is the corresponding author.}}
}
%

%
\institute{Jilin University \and
Engineering Research Center of Knowledge-Driven Human-Machine Intelligence, MOE\\
\email{\{jiangll21, chenpan21, yewang22\}@mails.jlu.edu.cn \\
\{wutr, ruim\}@jlu.edu.cn} }

\maketitle              
%
\begin{abstract}

Inferring missing regions from severely occluded point clouds is highly challenging.
Especially for 3D shapes with rich geometry and structure details, inherent ambiguities of the unknown parts are existing. 
Existing approaches either learn a one-to-one mapping in a supervised manner or train a generative model to synthesize the missing points for the completion of 3D point cloud shapes.
These methods, however, lack the controllability for the completion process and the results are either deterministic or exhibiting uncontrolled diversity.
Inspired by the prompt-driven data generation and editing, we propose a novel prompt-guided point cloud completion framework, coined P2M2-Net, to enable more controllable and more diverse shape completion.
Given an input partial point cloud and a text prompt describing the part-aware information such as semantics and structure of the missing region, our Transformer-based completion network can efficiently fuse the multimodal features and generate diverse results following the prompt guidance.
We train the P2M2-Net on a new large-scale PartNet-Prompt dataset and conduct extensive experiments on two challenging shape completion benchmarks.
Quantitative and qualitative results show the efficacy of incorporating prompts for more controllable part-aware point cloud completion and generation. 
Code and data are available at \url{https://github.com/JLU-ICL/P2M2-Net}.


\end{abstract}
\section{Introduction}
  \begin{figure*}[h]
  \centering
  \includegraphics[width=0.75\textwidth]{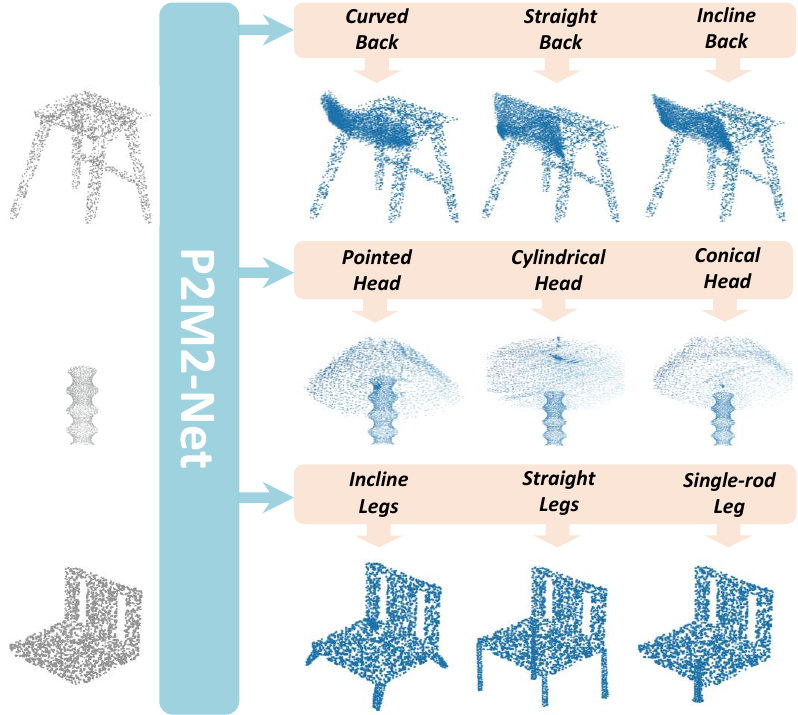}
  \caption{
  Given an input point cloud with a large missing region (left column), our P2M2-Net can use different text prompts to guide the shape completion and generate diverse outputs in a controllable manner. 
  }
  \label{fig:0}
\end{figure*}
Point cloud is one of the most commonly used 3D shape representations,
which requires less memory to store detailed geometry and structural information about a 3D shape.
Nowadays, point clouds can easily be obtained through depth cameras or other 3D scanning devices.
However, due to the resolution of the scanning devices, occlusion or the limitation of accessible scanning regions, raw point clouds are often sparse or incomplete.
Point cloud completion aims to completing the geometry and structure of the missing region given the partial point cloud as input.
When the missing region is significantly large, inherent \textit{ambiguities} may exist when performing the completion, i.e., there may be multiple options for the missing parts.
How to obtain the expected completion result is a challenging task for the point cloud completion.


Existing point completion methods \cite{yuan2018pcn,zhang2020pfnet,yu2021pointr,wen2021pmp,wen2022pmp,zhou2022seedformer,xiang2021snowflakenet} usually take the incomplete point cloud of a 3D shape as input and train an encoder-decoder network to map the input to a complete shape.
However, it is difficult to learn the mapping due to the sparsity and limited information from the input.
One way to provide more information for point cloud completion is using images as the guidance \cite{zhang2021view,garbade2019two,li2019rgbd,liu20203d,wang2022ffnet}.
Though ambiguities can be resolved and improved performance can be obtained by considering more image constraints during the completion, the images and point clouds need to be matched so that features from the image modality can be used to complete the missing features for the corresponding point clouds.
Furthermore, as no explicit semantic and structure information is considered in these methods, their completion mainly focus on the global geometry level.

On the other hand, the point cloud completion can also be regarded as a conditional generation problem, in which the complete shape is generated based on the input partial point cloud.
For example, some methods \cite{wu2020multimodal,arora2022multimodal,zhou20213d,achlioptas2018learning,zhang2021unsupervised} work on the multimodal shape completion with the goal of generating diverse 3D shapes from a single partial point cloud.
In this way, the point cloud completion is modeled as a one-to-many mapping which allows multiple outputs as long as the completed shapes respect to the input and their geometry and structure are plausible.
Although diverse results can be obtained from these generative approaches, it is hard to generate a specific complete shape that meets the expectation or requirement of the user. 
How to allow the user to control or guide the completion in an intuitive and efficient manner is worth to investigate.

In this paper, we aim for controllable point cloud completion that can accept a simple form of guidance (e.g., text prompt) to generate plausible outputs that satisfy the user's specification.
Also, in addition to the global geometry, we attempt to explicitly focus on the part semantics and structures when performing the completion.
Such part-aware modeling can allow more fine-grained control on the completion process.
To this end, we propose P2M2-Net, a novel \textbf{p}art-aware \textbf{p}rompt-guided \textbf{m}ultimodal point cloud completion framework, to enable more controllable and diverse shape completion.
With a text prompt describing the part-aware information such as semantics and structure of the missing region, the P2M2-Net can efficiently fuse the features from two modalities, i.e., 3D point clouds and text prompts, and predict a complete shape that matches to the text prompt.
The word \textit{multimodal} in our paper has two kinds of meanings: one indicates the completion is based on features from multiple data modalities; the other one represents that multiple different shapes can be generated when different text prompts are used as guidance for the same input point cloud (see Figure \ref{fig:0}).

To enable the joint learning between text prompts and the 3D point data, we construct a novel large-scale dataset, named PartNet-Prompt, which contains part-level text prompt annotations for three representative shape categories (chair, table and lamp) from the PartNet dataset \cite{mo2019partnet}.
Each prompt is a short text phrase that describes the geometry or structure of the corresponding part, such as \textit{inclined back}, \textit{straight legs} etc.
With the paired data from our PartNet-Prompt dataset, we first perform a cross-modal contrastive pre-training to align the part-level features of the text and 3D point.
For the prompt-guided completion network, we adopt a Transformer-based network PoinTr \cite{yu2021pointr} and adapt it to perform point cloud completion using multimodal features.
A new multimodal feature encoder is proposed to extract the feature of each modality and then fuse them together using a attention-based feature fusion module.
Next, the fused multimodal feature is passed to the multimodal query generator and the multimodal-based point cloud decoder to predict the complete shape in a coarse-to-fine manner.

To evaluate the performance of our P2M2-Net, we conduct extensive experiments on two challenging PartNet-based point cloud completion benchmarks.
Quantitative and qualitative comparisons with the state-of-the-art methods show the the efficacy of incorporating prompts for the guided completion.
We also perform ablation studies on the cross-modal pre-training and the attention-based multimodal feature fusion and the results verify the effectiveness of each module.

In summary, our contributions are as follows:
\begin{enumerate}
    \item We build the PartNet-Prompt, a novel large-scale dataset with part-level text prompt annotations. With the paired cross-modal data on the semantics and structures of shape parts, various applications such as part-aware point cloud completion and generation as well as fine-grained shape understanding and retrieval can be supported.  
    \item We propose P2M2-Net, a novel part-aware prompt-guided framework which can achieve the point cloud completion in a more controllable manner. A contrastive pre-training and a multimodal feature encoder are proposed to better align and fuse the cross-modal features.
    Once trained, when guided by different text prompts, the P2M2-Net can generate diverse results from a single input.
    \item Extensive experiments on two challenging PartNet-based shape completion benchmarks demonstrate the superiority of P2M2-Net comparing to the state-of-the-art point cloud completion methods.
    Also, our prompt-guided completion can also be regarded as cross-modal compositional modeling and the diverse results show its potential in generating novel shapes.
    
\end{enumerate}

\section{Related Work}
\subsection{3D Point Cloud Shape Completion}
Point cloud completion for 3D shapes has been widely studied in recent years.
The task is usually modeled as a one-to-one mapping which outputs a deterministic complete shape from a given input.
To learn the mapping, the input partial point cloud is often encoded into a feature vector using conventional point-based encoders such as \cite{qi2017pointnet,qi2017pointnet++,phan2018dgcnn}.
With the encoded point feature, PCN \cite{yuan2018pcn} designs a multi-stage decoder which first predict a coarse complete shape and then employs the FoldingNet \cite{yang2018foldingnet} to refine the initial result.
Furthermore, Transformer-based encoder-decoder \cite{yu2021pointr,zhou2022seedformer} which can learn more comprehensive relationships among the points has also been investigated.
Comparing to these methods, since we learn transferrable features via the cross-modal pre-training, we can enable one-to-many mapping by using different text prompt as guidance.

Meanwhile, some methods \cite{wu2020multimodal,arora2022multimodal,zhou20213d,achlioptas2018learning,zhang2021unsupervised} formulate the point cloud completion as a shape generation problem and employ generative models to obtain diverse completion results.
These methods can inherently learn a one-to-many mapping, but their completion is not controllable.
For example, the multimodal point cloud completion (or MPC) \cite{wu2020multimodal} develops the first shape completion framework which can generate multimodal (i.e., diverse) results based on the conditional generative modeling, but it is difficult to incorporate user's constraints into the completion process.
In contrast, our method allows intuitive user control in the form of text prompt and we can also generate diverse outputs when different text prompts are used.

\subsection{Multimodal-based Point Cloud Completion}
Since there are inherent ambiguities when completing the partial point cloud, information from other data modalities may be used to guide the completion process.
In ViPC \cite{zhang2021view}, a single-view image that matches to the target shape is used to provide the information about missing region.
The additional image information has also been explored in the completion of RGB-D scenes which contain severe missing data due to the occlusion.
With aligned RGB and depth images, some approaches \cite{garbade2019two,li2019rgbd,liu20203d,wang2022ffnet} propose different schemes to fuse the information from the multimodal input data.
Our method also takes the advantage of using multimodal input to alleviate the ambiguities for the completion.
Instead of the image, we utilize the text prompt which is more flexible to provide the shape completion guidance.
To resolve the domain gap between the text prompt and the 3D point cloud, we adopt the part-level cross-modal pre-training to obtain aligned and transferable features for each data modality.
Meanwhile, our multimodal Transformer can also efficiently fuse the text and point features and generate the completion result based on the multimodal input data.



\subsection{Prompt-Driven Multimodal Learning}
Recently, the prompt-driven multimodal learning has attracted great attention for its applications in zero-shot learning \cite{radford2021learning,jia2021scaling}, 2D/3D visual perception \cite{hegde2023clip,kirillov2023segment}, content generation \cite{ramesh2022hierarchical,sanghi2022clip,poole2022dreamfusion,nichol2022point,liu2022towards,bahmani2023cc3d} and editing \cite{li20223ddesigner,mikaeili2023sked,sella2023vox}.
With text or other types of the prompt \cite{zhang2023adding,mou2023t2i}, impressive 2D images and 3D models can be generated in a controllable manner.
One key for the success of prompt-driven learning is the large-scale pre-trained model such as CLIP \cite{radford2021learning} which learns aligned features from paired data of two modalities.
To enable the multimodal learning between the text prompt and 3D shapes, we construct PartNet-Prompt dataset by manually annotating the 3D parts of representative PartNet shapes with text descriptions about their geometry and structure.
With such part-level annotation, we can achieve more fine-grained control in prompt-guided part-aware completion and generation.

\subsection{Cross-Modal Contrastive Pre-Training}
Due to the difference between the data modalities, there is a large domain gap between the features of point clouds and text prompts.
To facilitate the multimodal feature fusion, the 3D point cloud and text features need to be aligned into the same embedding space.
Contrastive pre-training methods \cite{tian2020contrastive,desai2021virtex,morgado2021audio,liu2021contrastive,radford2021learning,jia2021scaling,JunnanLi2021AlignBF,yang2022vision,afham2022crosspoint,huang2022multi} has been widely used to learn aligned and transferrable features for data of different modalities, e.g., images and natural language.
For contrastive pre-training that involves 3D point clouds, CrossPoint \cite{afham2022crosspoint} jointly learns the aligned representations of 3D point cloud shapes and their corresponding images, while the learned representations are used for point cloud understanding tasks such as 3D classification and segmentation.
However, to the best our knowledge, contrastive pre-training has not been sufficiently explored for joint learning of point cloud and text features, nor the prompt-guided point cloud completion task.

\section{Method}
\begin{figure*}[t]
  \centering
  \includegraphics[width=1 \textwidth]{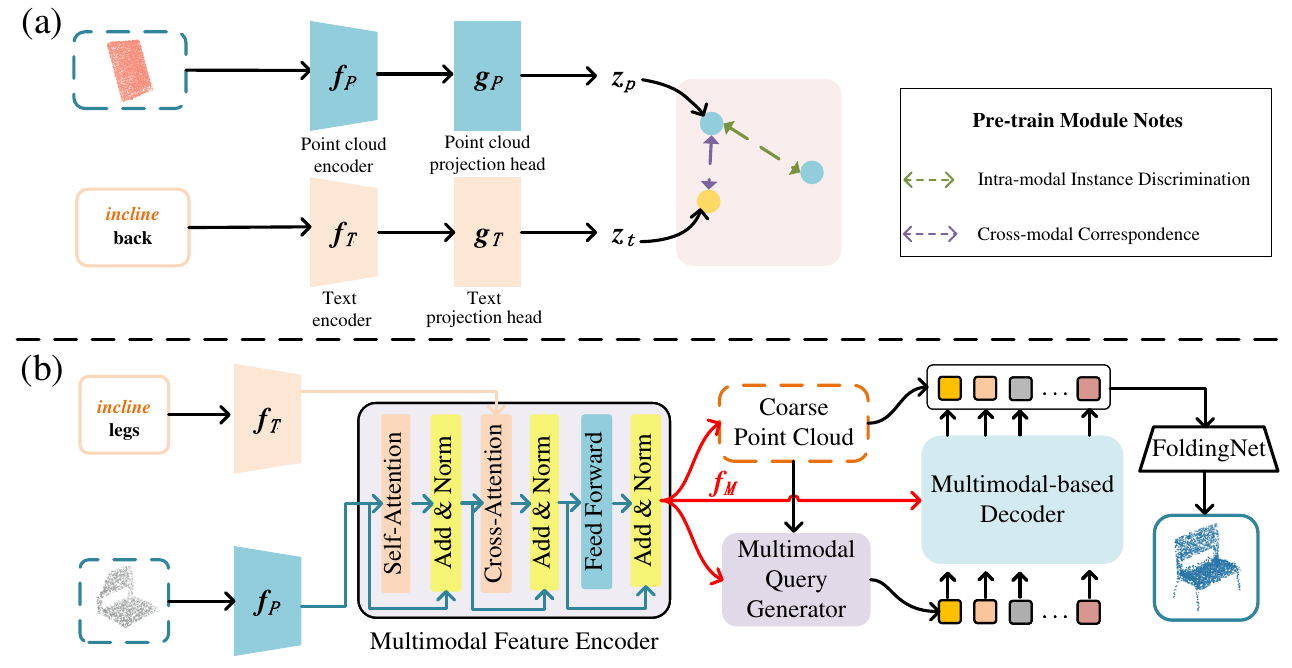}
  \caption{Overview of P2M2-Net. (a) Illustration of the cross-modal contrastive pre-training. Embeddings of the same part sare pulled closer; (b) Pipeline of the multimodal Transformer for prompt-guided completion.
  }
  \label{fig:2}
\end{figure*}
In this section, we introduce the PartNet-Prompt dataset and present the details of the P2M2-Net.
Figure \ref{fig:2} illustrates an overview of our pipeline which contains two stages of training: contrastive pre-training and multimodal Transformer for feature fusion and point cloud decoding.


\begin{table}[t]
\caption{The statistics of PartNet-Prompt dataset.  
}
\centering
\setlength{\tabcolsep}{2pt}
\resizebox{\linewidth}{!}{ 
\begin{tabular}{l|cccc|cc|ccc} 

\toprule
  Category       & \multicolumn{4}{c|}{Chair}      & \multicolumn{2}{c|}{Table}  & \multicolumn{3}{c}{Lamp}      \\ 
\hline
\#Shape    & \multicolumn{4}{c|}{8,176}      & \multicolumn{2}{c|}{9,906}  & \multicolumn{3}{c}{3,408~ ~}  \\
\# Part    & \multicolumn{4}{c|}{22,175}     & \multicolumn{2}{c|}{17,830} & \multicolumn{3}{c}{~9,201}    \\ 
\hline
Part-type      & Back  & Seat  & Armrest & Leg   & Tabletop & Leg              & Head  & Post  & Base          \\
\#Prompt-type   & 12    & 7     & 9       & 22    & 6        & 16               & 8     & 6     & 6             \\
\#Annotation        & 7,553 & 5,384 & 2,506   & 4,664 & 9,272    & 8,558            & 3,431 & 2,460 & 3,310         \\
\bottomrule
\end{tabular}}
\label{tab:dataset}
\end{table}
\subsection{PartNet-Prompt Dataset}
To enable part-level joint learning on the 3D shape and text prompt, we propose a large-scale PartNet-Prompt dataset which contains paired data of text prompt and their corresponding parts.
To construct the dataset, we manually annotate short text prompts for the pre-segmented parts of the Table, Chair, and Lamp categories based on their semantic segmentation from the original PartNet \cite{mo2019partnet} dataset.
A pre-defined set of vocabularies is combined with the semantic label of a part to describe part-level geometry, structure and semantics, e.g., curved back, single-rod leg, cylindrical head etc.
In total, the PartNet-Prompt contains part-level text prompt annotations for 8,176 chairs, 9,906 tables and 3,408 lamps, and the numbers of annotated parts are 22,175, 17,830 and 9,201, respectively.
The dataset is further split to train, validation and test set with the ratio 7:1:2 following the PartNet.
The statistics of the PartNet-Prompt dataset is shown in Table \ref{tab:dataset}.


\subsection{Contrastive Pre-Training for Prompt-Guided Completion}
Following the general idea of CrossPoint \cite{afham2022crosspoint}, we design a contrastive pre-training module that is specifically aiming for part-aware prompt-guided point cloud completion.
As shown in Figure \ref{fig:2}, this module contains a point cloud encoder and a text prompt encoder which can be some established networks, as well as two MLP-based projection heads to further map each modality into the target embedding space.
In the following, we introduce the details of the contrastive pre-training module.


\textbf{Point cloud encoder.}
We design an encoder $f_{P}$ that combines DGCNN \cite{phan2018dgcnn} and Point Transformer \cite{zhao2021point} to extract the point cloud feature. 
Given a point cloud w.r.t. a particular shape part, DGCNN is used to extract the local features for a set of points sampled by farthest point sampling (FPS).
Then, the global point cloud feature is obtained using Point Transformer in which self-attention is applied to mine the relationships among the local features.  


\textbf{Text prompt encoder.}
Given a text prompt that describes the geometry, structure and semantics of the corresponding part, we apply BERT \cite{devlin2018bert} as the text prompt encoder $f_{T}$ to extract a 768-dim vector as the initial feature for the text prompt.


\textbf{Contrastive pre-training of point cloud and text prompt.} 
With the initial features extracted by point cloud encoder $f_{P}$ and text prompt encoder $f_{T}$, the point cloud projection head $g_P$ and text projection head $g_T$ further map the features into a unified embedding space ${R}^{256}$.
For a shape part represented by point cloud $p_i$ and text prompt $t_i$, we denote the final feature embedding of each modality as  $z_P^i=g_{P}^i(f_{P}^i(p_i))$ and $z_T^i=g_{T}^i(f_{T}^i(t_i))$.
To align the point cloud and text prompt features, contrastive pre-training adopts the InfoNCE loss \cite{AaronvandenOord2018RepresentationLW} to fine-tune the encoders $f_{P}$, $f_{T}$ and projection heads $g_P$ and $g_T$.
Formally, the InfoNCE loss for contrastive pre-training is defined as:
\begin{equation}
L_{InfoNCE}=(L_{P\rightarrow T} + L_{T\rightarrow P})/2, \quad where
\end{equation}
\begin{equation}
L_{P\rightarrow T} = -\frac{1}{N}\sum_{i=1}^{N}log\frac{exp(s(z_P^i, z_T^i)/\tau)}{\sum_{k=1}^{N}exp(s(z_P^i, z_T^k)/\tau)},
\end{equation}
and $L_{T\rightarrow P}$ is defined similarly as $L_{P\rightarrow T}$.
Here, $s(\cdot,\cdot)$ is the cosine similarity between features of cross-modal pairs and $\tau$ is a temperature parameter which can adjust the feature learning.
By performing the cross-modal pre-training using $L_{InfoNCE}$, the final embeddings of the same shape part, e.g., $z_P^i$ and $z_T^i$ are pulled closer, and the embeddings of the different parts, e.g., $z_P^i$ and $z_T^k$, (when $i\neq k$), are repelled far away from each other.
Similar to CrossPoint \cite{afham2022crosspoint}, after pre-traning, the features extracted by the encoders $f_{P}$ and $f_{T}$ are passed to the downstream point cloud completion task.

\subsection{Multimodal Transformer for Point Cloud Completion}
To achieve the prompt-guided point cloud completion, we extend the PoinTr \cite{yu2021pointr} to work with multimodal features.

\textbf{Multimodal feature encoder.}
With the features extracted by the point cloud and text prompt encoders, we design a multimodal feature fusion module which contains self-attention and cross-attention layers as in \cite{vaswani2017attention} to fuse the features into a 1024-dim multimodal feature $f_M$.
Moreover, a coarse point cloud which can be used to guide the following completion process is also generated by applying two additional linear layers to $f_M$ and reshaping the output.

\textbf{Multimodal query generator.}
The query generator, which is composed of three Conv1D layers, takes the fused multimodal feature $f_M$ and coarse point cloud as input and generate a sequence of multimodal query proxies which can be used to query the related features from $f_M$.
Similar to PointTr \cite{yu2021pointr}, the coarse point cloud is utilized to provide the spatial coordinates when generating the query proxies.

\textbf{Multimodal-based point cloud decoder.}
With the multimodal query proxies and the fused multimodal feature $f_M$ as input, the point cloud decoder first predicts a sequence of predicted proxies which contain the multimodal information.
When performing the querying, a kNN model is used to query the multimodal features around each point in the coarse point cloud.
Then, the FoldingNet \cite{yang2018foldingnet} is employed to recover detailed local shapes centered around the generated proxies. 
Note that we only predict the missing part that matches to the text prompt and concatenate the output to the input point cloud to obtain a complete shape. 

\textbf{Training Details.}
The P2M2-Net is trained in two stages using the annotated data from the PartNet-Prompt dataset.
Each part is uniformly sampled into 1024 points similar to MPC \cite{wu2020multimodal}.
Separate encoders or models are trained during the contrastive pre-training stage and the prompt-guided completion stage, using the parts from the training set of Chair, Table and Lamp, respectively.
For the contrastive pre-training stage, the InfoNCE loss $L_{InfoNCE}$ is used and we train the point cloud encoder from scratch and the text prompt encoder based on the pre-trained BERT.
The pre-training is running for 2000 epoches.
For the prompt-guided completion stage, we use the Chamfer Distance (CD) loss and train the multimodal Transformer-based network for 500 epoches.

\section{Experiments}
We conduct quantitative and qualitative evaluations of P2M2-Net on the prompt-eguided completion.
We also perform ablation studies to verify the effectiveness of key modules in P2M2-Net.



\subsection{Evaluation Metrics and Benchmarks}   
Following previous works \cite{yuan2018pcn,wu2020multimodal,yu2021pointr}, we adopt the following metrics for quantitative evaluation of the completion or diverse generation performance:

\textbf{Chamfer Distance (CD)} \cite{fan2017point}: 
  the average Chamfer Distance which measures the set-wise distance between the completed and the ground truth point clouds is used as a metric for evaluating the prompt-guided completion.


  \textbf{F1 Score (F-Score)} \cite{tatarchenko2019single}:
  F1 score is defined as the harmonic mean of precision and recall when performing 3D classification, reconstruction or completion.
  It explicitly evaluates the distance between the predicted points and the GT points and intuitively measures the percentage of points that is predicted correctly.
  We set the distance threshold $d=0.01$ when computing the F1 score.
  
  
  \textbf{Total Mutual Difference (TMD)} \cite{wu2020multimodal} : 
  given $k$ completion results for an input, the mutual difference $d_{i}$ is defined as the average Chamfer Distance between the i-th shape to the other $k-1$ shapes.
  Then, the Total Mutual Difference is define as $\Sigma _{i=1}^{k}d_{i}$ to measure the \textbf{diversity} of the completion results.
  
  \textbf{Minimal Matching Distance (MMD)} \cite{Achlioptas_Diamanti_Mitliagkas_Guibas_2017} :
  for each input, we calculate the minimal matching distance between 10 predicted shapes w.r.t the corresponding ground truth shape to measure the \textit{quality} of the completion results.
  
  \textbf{Unidirectional Hausdorff Distance (UHD)} \cite{wu2020multimodal}: 
  we compute the average Hausdorff distance from the input partial point cloud to each of the completed shapes to measure the \textit{fidelity} of the completion results.
\textbf{Benchmarks for Point Completion.}
We follow the settings in MPC \cite{wu2020multimodal} and use their released code to generate two challenging benchmarks to evaluate the P2M2-Net for part-ware completion.
The \textit{PartNet} benchmark is obtained by removing points of randomly selected parts from the shapes in the PartNet dataset \cite{mo2019partnet}.
The \textit{PartNet-Scan} benchmark is obtained by first randomly removing the parts and then virtually scanning the remaining parts.
These two benchmarks both focus on part-level incompleteness and exhibit high ambiguities for the missing region.


\subsection{Evaluation for One-to-One Completion}
We first conduct comparisons following the conventional setting in previous methods, i.e., predicting a complete shape given a partial input.
For our method, since the text prompt is needed to provide the guidance for completion, we use GT text prompt of the missing part as an additional input.
At the first glance, using the GT text prompt to provide additional information seems to be unfair for other methods when conducting the comparison.
Meanwhile, since our method can be regarded as a multimodal version of PoinTr \cite{yu2021pointr}, the comparison still can show how much improvement can be achieved by incorporating the prompt guidance comparing the the PoinTr baseline and other state-of-the-art methods.

\textbf{Quantitative comparison.}
For each benchmark, we compare our results with representative and state-of-the-art point cloud completion methods, i.e., PCN \cite{yuan2018pcn}, PFNet \cite{zhang2020pfnet} , FoldingNet \cite{yang2018foldingnet}, MPC \cite{wu2020multimodal}, TopNet \cite{tchapmi2019topnet}, PoinTr \cite{yu2021pointr}, and PMP-Net++ \cite{wen2022pmp}.
For each compared method, we use their released code and train their models on the training set of our benchmark.
Table \ref{tab:partnet} and \ref{tab:partnetscan} show the results of quantitative comparison measured by CD and F-Score.
It can be observed that for both benckmarks, our P2M2-Net outperforms most of the baseline methods and just underperforms in a few cases comparing to PoinTr and PMP-Net++.
The reasons may be as follows:
1) by fusing the text prompt feature to the point cloud feature, some noise may actually be introduced since the pre-training is not perfect due to the amount of data and the ambiguity between different parts;
2) since the same text prompt may be used to annotate parts with minorly different geometry, the learned text feature may be related to an average shape of the described part.
Nevertheless, by incorporating with the text prompt, our method can guide the completion with the expected geometry and structure, while makes the completion more controllable.

\begin{table}[t]
\caption{Quantitative comparison on the PartNet benchmark. CD-L2($\times10^3$) and F-Score@0.01 are used to compare with other methods.
}
\centering
\setlength{\tabcolsep}{2pt}
\resizebox{0.8\linewidth}{!}{ 
\begin{tabular}{l|cl|ll|ll} 
\toprule
\multicolumn{1}{c|}{\multirow{2}{*}{Method}} & \multicolumn{2}{c|}{Chair}                    & \multicolumn{2}{c|}{Table}                              & \multicolumn{2}{c}{Lamp}                                \\ 
\cline{2-7}
\multicolumn{1}{c|}{}                        & $CD$           & \multicolumn{1}{c|}{F-Score} & \multicolumn{1}{c}{$CD$} & \multicolumn{1}{c|}{F-Score} & \multicolumn{1}{c}{$CD$} & \multicolumn{1}{c}{F-Score}  \\ 
\hline
PCN \cite{yuan2018pcn}                                         & 2.098          & 0.152                        &  3.560                   &  0.131                       & 9.133                    & 0.110                        \\
PFNet  \cite{zhang2020pfnet}                                     & 3.734          & 0.087                        &  6.282                   &  0.120                       & 14.652                   & 0.075                        \\
FoldingNet  \cite{yang2018foldingnet}                                 & 2.733          & 0.082                        &  5.194                   &  0.193                       & 12.466                   & 0.116                        \\
MPC     \cite{wu2020multimodal}                                    & 2.081          & 0.132                        &  4.132                   &  0.236                       & 10.465                   & 0.088                        \\
TopNet \cite{tchapmi2019topnet}            & 1.480          & 0.171                        &  3.069                   &  0.174                       & 7.388                     & 0.081                        \\
PoinTr  \cite{yu2021pointr}                                     & 1.292          & 0.364                        &  2.682                   &  0.356                       &  6.017                  & 0.354                        \\
PMP-Net++  \cite{wen2022pmp}                                  & \textbf{1.236} & \textbf{0.385}               &  2.427                   &  0.369                       & 5.987                    & 0.326                        \\
\hline
P2M2-Net                                        & 1.351          & 0.333                        &  \textbf{2.320 }         &  \textbf{0.373}              & \textbf{5.675}           & \textbf{0.372}               \\
\bottomrule
\end{tabular}}
\label{tab:partnet}
\end{table}

\begin{table}[t]
\caption{Quantitative comparison on the PartNet-Scan benchmark. CD-L2($\times10^3$) and F-Score@0.01 are used to compare with other methods.
}
\centering
\setlength{\tabcolsep}{2pt}
\resizebox{0.8\linewidth}{!}{ 
\begin{tabular}{l|cl|ll|ll} 
\toprule
\multicolumn{1}{c|}{\multirow{2}{*}{Method}} & \multicolumn{2}{c|}{Chair}                    & \multicolumn{2}{c|}{Table}                              & \multicolumn{2}{c}{Lamp}                                \\ 
\cline{2-7}
\multicolumn{1}{c|}{}                        & $CD$           & \multicolumn{1}{c|}{F-Score} & \multicolumn{1}{c}{$CD$} & \multicolumn{1}{c|}{F-Score} & \multicolumn{1}{c}{$CD$} & \multicolumn{1}{c}{F-Score}  \\ 
\hline
PCN      \cite{yuan2018pcn}                                    & 3.421          &  0.097                      &  4.662                   &  0.104                    & 10.019                    &  0.091                   \\
PFNet    \cite{zhang2020pfnet}                                 & 4.571          &  0.065                      &  7.031                   &  0.098                    & 15.351                    &  0.073                    \\
FoldingNet   \cite{yang2018foldingnet}                         & 3.843          &  0.071                      &  5.972                   &  0.120                    & 13.544                    &  0.111                    \\
MPC    \cite{wu2020multimodal}                                 & 3.464          &  0.074                      &  4.694                   &  0.213                    & 11.096                    &  0.076                    \\
TopNet  \cite{tchapmi2019topnet} & 2.016          &  0.117                      &  3.473                   &  0.120                    & 6.972                     &  0.079                     \\
PoinTr   \cite{yu2021pointr}                                   & 1.325          &  0.359                      &  2.990                   &  0.343                    & 7.623                     &  0.301                     \\
PMP-Net++   \cite{wen2022pmp}                                  & \textbf{1.294} &  \textbf{0.375}             &  2.641                   &  0.357                    & \textbf{6.132}            &  \textbf{0.318 }                      \\
\hline
P2M2-Net                                                          & 1.421          & 0.356                       &  \textbf{2.423}          &  \textbf{0.361}            & 6.307                    &  0.305             \\
\bottomrule
\end{tabular}}
\label{tab:partnetscan}
\end{table}


\begin{figure}[t]
  \centering
  \includegraphics[width=1 \textwidth]{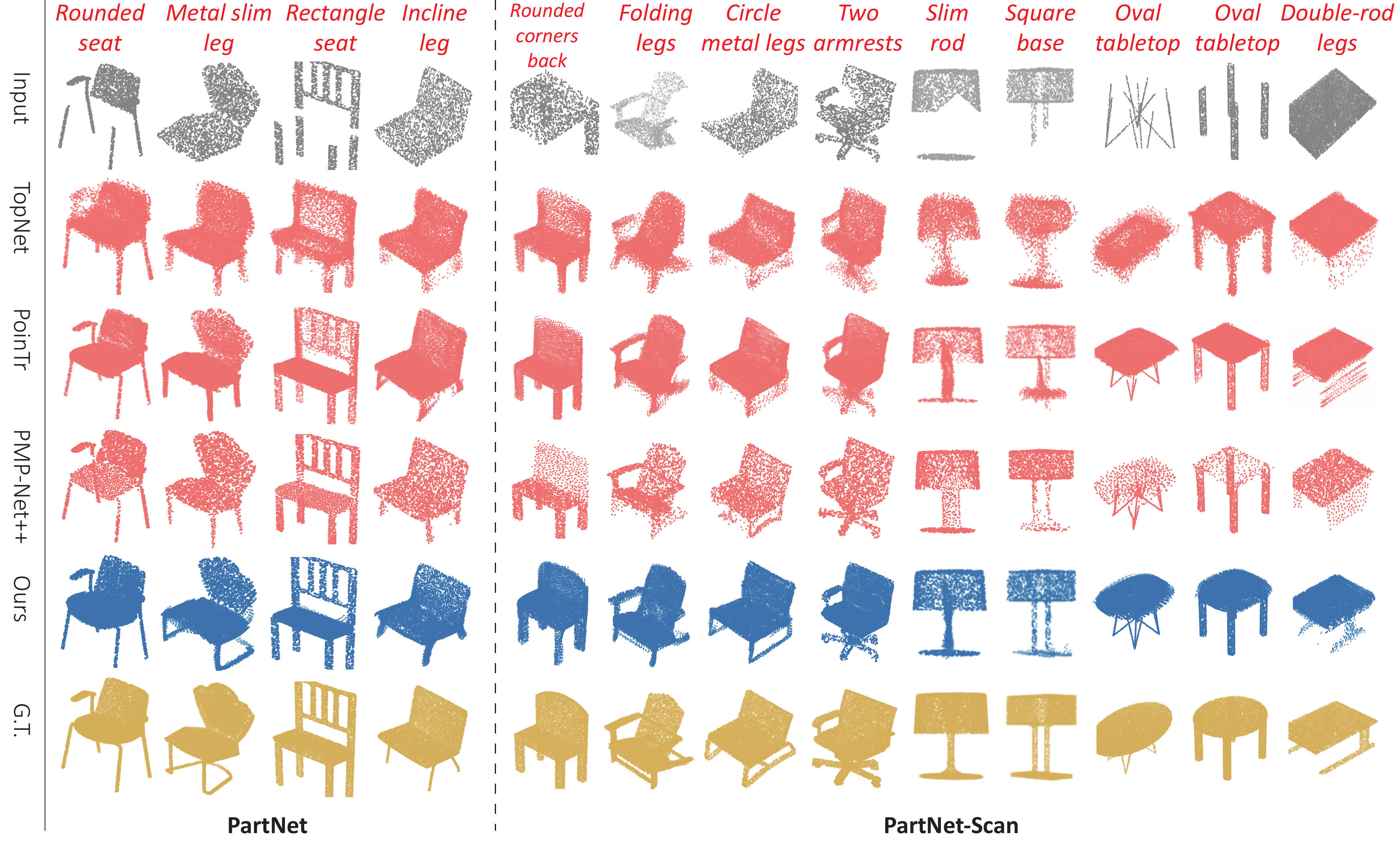}
  \caption{Qualitative comparisons on PartNet and PartNet-Scan benchmarks. 
  }
  \label{fig:3}
\end{figure}

\textbf{Qualitative comparison.}
Figure \ref{fig:3} shows the qualitative results of methods which achieve relative high performance in the quantitative comparison, i.e., TopNet\cite{tchapmi2019topnet}, PoinTr\cite{yu2021pointr} and PMP-Net++\cite{wen2022pmp}, on two benchmark datasets, namely PartNet (dashed line on the left) and PartNet-scan (dashed line on the right).
Among the compared methods, PMP-Net++ achieves the best performance in terms of the quality of the results and the similarity to the GT.
However, all methods including PMP-Net++ cannot predict the expected shape for certain cases, such as the chair leg in the second column and the back with rounded corners in the fifth column of Figure \ref{fig:3}.
In contrast, our method can generate the expected shape when the text prompt is used as guidance.

\begin{figure}[t]
  \centering
  \includegraphics[width=0.81 \textwidth]{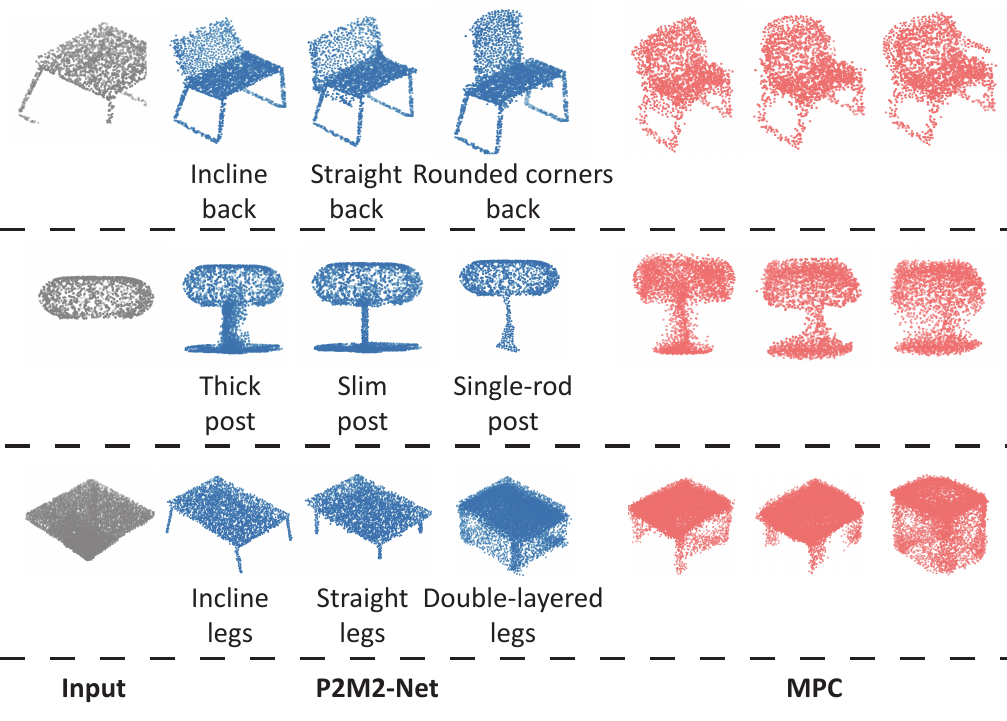}
  \caption{
  Qualitative comparison with MPC \cite{wu2020multimodal} for shapes in the PartNet benchmark.
  }
  \label{fig:compare with mpc}
\end{figure}

\begin{figure*}[!h]
  \centering
  \includegraphics[width=1\textwidth]{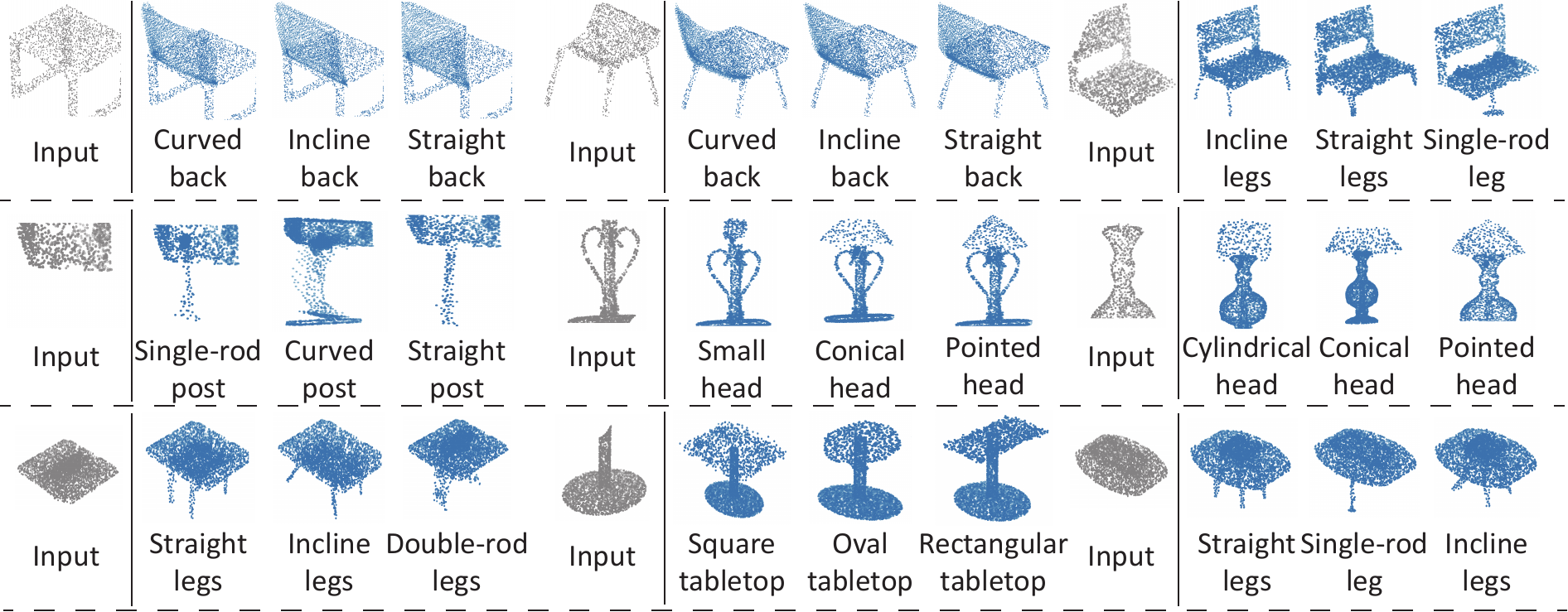}
  \caption{Qualitative results on PartNet and PartNet-Scan. We visualize the generated results from different prompts. P2M2-Net not only preserves the originally observed structure but also achieves diverse generated results that comply with the prompt. }
  \label{fig:diversity}
\end{figure*}

\subsection{Evaluation for Multimodal (Diverse) Completion}

With different text prompts as input, our P2M2-Net can also generate multimodal (or diverse) completion results.
Table \ref{tab:diversity} shows the quantitative comparisons with MPC \cite{wu2020multimodal} and the baselines used in their paper.
Since we use the same benckmarks \textit{PartNet} and \textit{PartNet-Scan} as MPC, we directly compare with the metrics reported in their paper.
For other methods, we randomly select one text prompt from the set of prompt annotations for the corresponding part type and generate the result with the prompt-guidance.
From the results, our method achieves the best MMD (quality), TMD (diversity) and UHD (fidelity) metrics.
Figure \ref{fig:compare with mpc} shows qualitative comparisons with MPC and our results achieve both superior diversity and quality.
In Figure \ref{fig:diversity}, more qualitative results are provided.
It can be seen our method can generate diverse, plausible or even novel shapes with different prompts.

\subsection{Ablation Studies}

\label{sec:Ablation}



\begin{table}[t]
\caption{Quantitative comparison for multimodal completion on PartNet benchmark. Note that MMD (quality), TMD (diversity) and UHD (fidelity) are multiplied by $10^{3}$, $10^{2}$ and $10^{2}$, respectively.}
\setlength{\tabcolsep}{2pt}
\centering
\resizebox{0.9\linewidth}{!}{ 
\centering
\begin{tabular}{l|cll|cll|cll} 
\toprule
PartNet     & \multicolumn{3}{c|}{MMD↓}                                                           & \multicolumn{3}{c|}{TMD↑}                                                           & \multicolumn{3}{c}{UHD↓}                                                            \\ 
\hline
Method      & \multicolumn{1}{c|}{Chair} & \multicolumn{1}{c|}{Table} & \multicolumn{1}{c|}{Lamp} & \multicolumn{1}{c|}{Chair} & \multicolumn{1}{c|}{Table} & \multicolumn{1}{c|}{Lamp} & \multicolumn{1}{c|}{Chair} & \multicolumn{1}{c|}{Table} & \multicolumn{1}{c}{Lamp}  \\ 
\hline
pcl2pcl \cite{Chen_Chen_Mitra_2020}     & 1.90                       & 1.90                       & 2.50                      & 0.00                       & 0.00                        & 0.00                       & 4.88                       & 4.64                       & 4.78                      \\
KNN-latent  & 1.39                       & 1.30                       & 1.72                      & 2.28                       & 2.36                       & 4.18                      & 8.58                       & 7.61                       & 8.47                      \\
MPC \cite{wu2020multimodal}      & 1.52  & 1.46                       & 1.97                      & 2.75                       & 3.30                       & 3.31                      & 6.89                       & 5.56                       & 5.72                      \\
\hline
P2M2-Net     & \textbf{1.35}             & \textbf{1.39}                  & \textbf{1.62}             & \textbf{3.07}              & \textbf{3.51}             & \textbf{3.41}               & \textbf{2.65}       &  \textbf{2.53}           & \textbf{3.24}                    \\

\bottomrule
\end{tabular}}
\label{tab:diversity}
\end{table}

\begin{table}[t]
\caption{Quantitative ablation study on the PartNet benchmark.
The effectiveness of cross-modal pre-training (Pre-train) and attention-based feature fusion (Attention) are evaluated.
We investigate different designs including Pre-train Module (Pre-train) and Multi-modal Fusion Module (Attention)}
\setlength{\tabcolsep}{2pt}
\resizebox{\linewidth}{!}{ 
\centering
\begin{tabular}{l|cc|lll|lll|lll} 
\toprule
\multirow{2}{*}{Model} & \multicolumn{1}{l}{\multirow{2}{*}{Pre-train}} & \multicolumn{1}{l|}{\multirow{2}{*}{Attention}} & \multicolumn{3}{c|}{Chair} & \multicolumn{3}{c|}{Table} & \multicolumn{3}{c}{Lamp}  \\ 
\cline{4-12}
                       & \multicolumn{1}{l}{}                           & \multicolumn{1}{l|}{}                                 & CD    & TMD  & UHD         & CD    & TMD  & UHD         & CD    & TMD  & UHD        \\ 
\hline
A                      &                                                &                                                       & 1.669 & 0.26 & 4.58        & 2.714 & 0.19 & 4.37        & 6.208 & 0.22 & 4.63       \\
B                      &     \checkmark                                 &                                                       & 1.425 & 1.32 & 3.42        & 2.503 & 1.29 & 2.98        & 5.864 & 1.80 & 4.05       \\
P2M2-Net                      &     \checkmark                                 &   \checkmark                                          & \textbf{1.365} & \textbf{3.07} & \textbf{2.65}        & \textbf{2.320} & \textbf{3.51} & \textbf{2.53}        & \textbf{5.675} & \textbf{3.41} & \textbf{3.24}       \\
\bottomrule
\end{tabular}}
\label{tab:Ablation}
\end{table}

\begin{figure}[t]
  \centering
  \includegraphics[width=0.7 \textwidth]{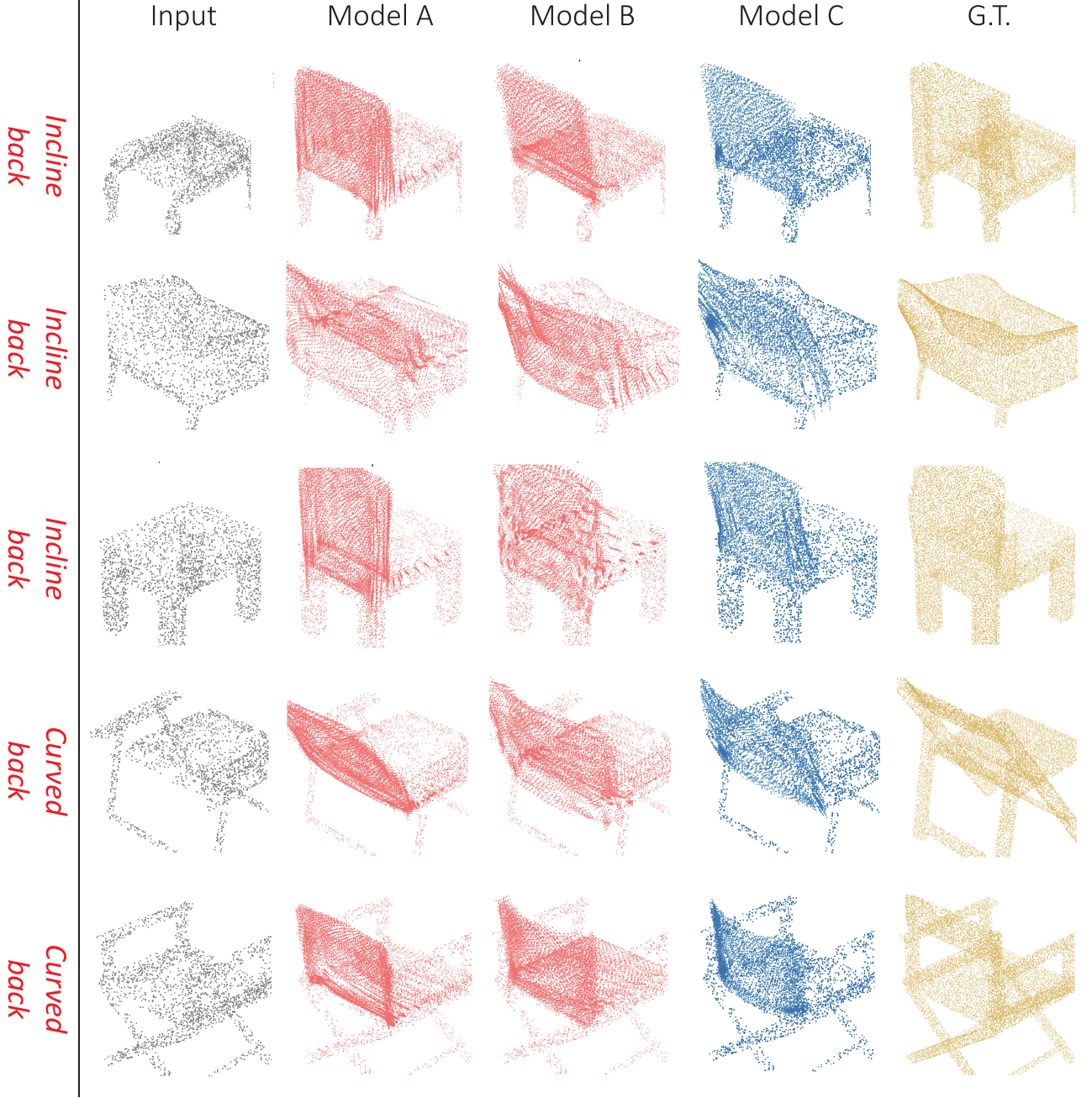}
  \caption{
  Qualitative ablation study on cross-modal pre-training and attention-based feature fusion modules. 
  }
  \label{fig:ablation}
\end{figure}

We conduct ablation studies to evaluate the key modules of our framework: cross-modal pre-training and attention-based feature fusion.
We implement two variations of methods that have the corresponding module disabled.
The baseline model A directly uses features from pre-trained models of DGCNN \cite{phan2018dgcnn} and BERT, and performs the feature fusion by simple concatenation.
Such model A can be regarded as a simple multimodal extension of PoinTr \cite{yu2021pointr}.
The baseline model B performs the cross-modal pre-training but still uses the simple concatenation-based feature fusion.
Table \ref{tab:Ablation} and Figure \ref{fig:ablation} show the quantitative and qualitative results of ablation studies.
It can be seen both the two modules are important to achieve diverse results while respecting to the input prompts.

\begin{figure}[!h]
  \centering
  \includegraphics[width=0.85\textwidth]{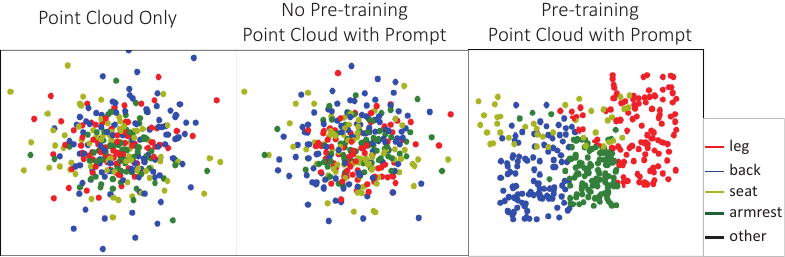}
  \caption{t-SNE visualization of features obtained with different schemes (see the main text for more details).
  }
  \label{fig:tsne}
\end{figure}

To further examine the effectiveness of the cross-modal pre-training, we use t-SNE to visualize the embedding space of different features for 150 chairs with 394 parts in Figure \ref{fig:tsne}.
It can be seen the initial point cloud features of different parts are mixed together.
This is because different parts may have similar shapes.
If simply concatenating the point cloud feature with the text prompt feature without pre-training, the fused features are still not representative to corresponding parts.
After performing pre-training, since the point cloud and text prompt are aligned into the same space, even if we just simply concatenate them together, the resulting features can better represent the parts as the parts with similar geometry and semantics are closer in the space.




\section{Conclusion}

In this paper, we propose P2M2-Net, a novel part-aware prompt-guided framework for multimodal point cloud completion.
With the guidance of the text prompt, our P2M2-Net can resolve the ambiguities for the large missing region and enable more controllable completion process.
Moreover, with different text prompts as input, we can also generate diverse completion results for the same partial point cloud.
To enable the joint learning of point cloud and text prompts, we construct a novel large-scale dataset PartNet-Prompt which has the potential to support more part-level multimodal learning tasks such as prompt-guided generation and editing.
Currently, our P2M2-Net is based on supervised learning with the paired data from PartNet-Prompt and our completion is still deterministic when the same text prompt is used to the same partial point cloud.
In the future, we would like to introduce generative models such as GAN or diffusion models to achieve more diverse results for controllable part-aware shape completion and generation.

\bibliographystyle{splncs04}
%
\bibliography{refs}





\end{document}